\def\year{2022}\relax
\documentclass[letterpaper]{article} 
\usepackage{aaai22}  
\usepackage{times}  
\usepackage{helvet}  
\usepackage{courier}  
\usepackage[hyphens]{url}  
\usepackage{graphicx} 
\usepackage{color}
\usepackage{natbib}  
\usepackage{caption} 
\DeclareCaptionStyle{ruled}{labelfont=normalfont,labelsep=colon,strut=off} 
\usepackage{algorithm}
\usepackage{algorithmic}

\usepackage{dingbat}
\usepackage{newfloat}
\usepackage{listings}
\floatstyle{ruled}
\newfloat{listing}{tb}{lst}{}
\floatname{listing}{Listing}
\title{Sparse MLP for Image Recognition: Is Self-Attention Really Necessary?}

\author{
    
    Chuanxin Tang$^{1}$\thanks{Equal contribution},
    Yucheng Zhao$^{2}$\footnotemark[1]\thanks{Interns at MSRA},
    Guangting Wang$^{2}$\footnotemark[1]\footnotemark[2],
    Chong Luo$^{1}$, \\
    Wenxuan Xie$^{1}$,
    Wenjun Zeng$^{1}$
}

\affiliations{
    \textsuperscript{\rm 1}Microsoft Research Asia, Beijing, China\\
    \textsuperscript{\rm 2}University of Science and Technology of China, Hefei, China\\

    \{chutan, cluo, wenxie, wezeng\}@microsoft.com,\{lnc, flylight\}@mail.ustc.edu.cn
}

\usepackage{bibentry}

\begin{document}

\maketitle

\begin{abstract}
Transformers have sprung up in the field of computer vision. In this work, we explore whether the core self-attention module in Transformer is the key to achieving excellent performance in image recognition. To this end, we build an attention-free network called sMLPNet based on the existing MLP-based vision models. Specifically, we replace the MLP module in the token-mixing step with a novel sparse MLP (sMLP) module. For 2D image tokens, sMLP applies 1D MLP along the axial directions and the parameters are shared among rows or columns. By sparse connection and weight sharing, sMLP module significantly reduces the number of model parameters and computational complexity, avoiding the common over-fitting problem that plagues the performance of MLP-like models. When only trained on the ImageNet-1K dataset, the proposed sMLPNet achieves 81.9\% top-1 accuracy with only 24M parameters, which is much better than most CNNs and vision Transformers under the same model size constraint. When scaling up to 66M parameters, sMLPNet achieves 83.4\% top-1 accuracy, which is on par with the state-of-the-art Swin Transformer. The success of sMLPNet suggests that the self-attention mechanism is not necessarily a
silver bullet in computer vision. The code and models are publicly available at {\footnotesize\color{black}{\url{https://github.com/microsoft/SPACH}}}.
\end{abstract}




\section{Introduction}
Since the proposal of AlexNet \cite{krizhevsky2012imagenet}, convolutional neural networks (CNNs) have been the dominant design paradigm in computer vision. 
This situation has recently been changed by the boom of conv-free vision Transformers. The pioneering work ViT \cite{dosovitskiy2020image} interprets an image as a sequence of patches and processes it by a standard Transformer encoder as used in natural-language processing (NLP). Specifically, an image is divided into non-overlapping patches, and the sequence of linear embeddings of these patches are used as an input to the vision Transformer. The encoding process involves alternate spatial mixing and channel mixing modules implemented by multi-head self-attention and feed forward networks (FFNs), respectively. ViT performs very well on image recognition tasks when pretrained on a very large dataset. Shortly after, DeiT \cite{touvron2021training} further demonstrates that a conv-free vision Transformer can achieve state-of-the-art (SOTA) image recognition accuracy when pretrained on ImageNet-1K, with appropriate data augmentation and model regularization techniques.

The conv-free vision transformers are actually promoting two views. First, global dependency modeling is important. Not only that, it could even completely replace local dependency modeling which used to be baked into the model by convolutions. Second, self-attention is important. Despite the good performance of ViT and DeiT, it is clear that academia has not fully embraced both views, and questions about them are often heard. 

\begin{figure}[t]
\centering
\includegraphics[width=0.95\columnwidth]{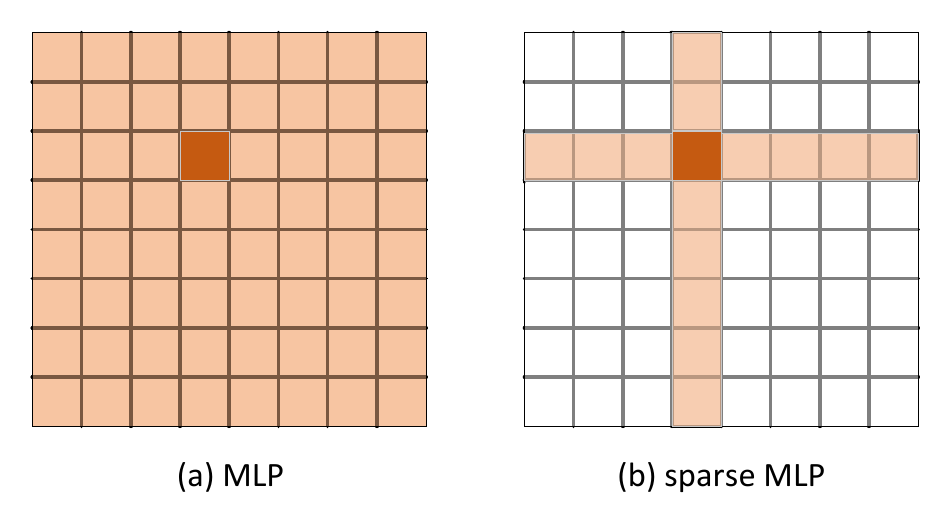} 
\caption{The proposed sparse MLP reduces the computational complexity of MLP by sparse connection and weight sharing. In MLP (a), the token in dark orange interacts with all the other tokens in a single MLP layer. In contrast, in one sMLP layer (b), the dark-orange token only interacts with horizontal and vertical tokens marked in light orange. The interaction with all the other white tokens can be achieved when sMLP is executed twice.}
\label{fig:sMLPinteraction}
\end{figure}

On the one hand, researchers challenge the complete disposal of locality bias. It is natural to ask, since locality is always valid in natural images, why bother learning it through global self-attention modules instead of directly injecting it into the network? Besides, global self-attention has quadratic computational complexity with respect to the number of input tokens. As a result, the network structure does not favor high-resolution input and is not friendly to pyramid structure. These are considered drawbacks of vision Transformers as high-resolution input and pyramid structure have been widely acknowledged to improve the image recognition accuracy. The recent Swin Transformer \cite{liu2021swin} injects locality back into the network by limiting self-attention operations within a local window. This setting also controls the computational complexity and allows for a pyramid structure or multi-stage processing. The superior performance of Swin Transformer demonstrates the value of locality bias and multi-stage processing. 

On the other hand, researchers also challenge the need for self-attention. MLP-Mixer \cite{tolstikhin2021mlp} recognizes the importance of modeling global dependencies, but it adopts an MLP block, instead of a self-attention module, to achieve it. The overall architecture of MLP-Mixer is similar to ViT. An input image is divided into patches which are then mapped into tokens. The encoder also contains alternating layers for spatial mixing and channel mixing. The only major difference is that the spatial mixing module is implemented by an MLP block. MLP-Mixer inherits all the drawbacks of ViT, besides, it is prone to over-fitting due to the excessive number of parameters. It is not surprising that there is still an accuracy gap between MLP-Mixer and SOTA models, especially on mid-scale datasets such as ImageNet. However, MLP-Mixer's defeat cannot convince us of the need for self-attention. 
We ask: is it possible for an attention-free network to achieve SOTA performance on image recognition after addressing all the drawbacks?


The work to be presented in this paper confirms that the answer to the above question is yes. We design an attention-free network, called sMLPNet, which only uses convolution and MLP as building blocks. sMLPNet adopts a similar architecture as ViT and MLP-Mixer, and the channel mixing module is exactly the same. In each token mixing module, depth-wise convolution is adopted to take advantage of the locality bias and a modified MLP is used to model global dependencies. Specifically, we propose sparse-MLP (sMLP) module which is featured by axial global dependency modeling, as shown in Fig.\ref{fig:sMLPinteraction}, and weight sharing. sMLP significantly reduces the computational complexity and allows us to adopt a pyramid structure for multi-stage processing. As a result, the sMLPNet is capable of achieving top image recognition performance on par with Swin Transformer at an even smaller model size. 

In a nutshell, in the craze of vision Transformers, we investigate whether the key component of Transformers, known as self-attention, is the true game-changing factor for image understanding. Based on the learning from past vision models, we retain design ideas that are important to image understanding, such as locality and pyramid structure. We also absorb the idea of global dependency modeling, but choose to implement it with a sparse MLP module we propose. We eventually build an attention-free network called sMLPNet which achieves SOTA image recognition performance. Our work suggests that self-attention might not be a core component in vision model design. Instead, proper use of locality, pyramid structure and careful control of computational complexity are the keys to designing a high-performance vision model.

\section{Related Work}
\subsection{CNN-Based Vision Models}
Since the success of AlexNet \cite{krizhevsky2012imagenet},
CNNs have been the mainstream tools in computer vision field. VGG \cite{simonyan2014very} demonstrated that a series of convolutions with a small 3x3 receptive field is sufficient to train state-of-the-art models. Later, ResNet \cite{he2016deep} introduced skip-connections together with the batch normalization layer \cite{ioffe2015batch}, which enabled training of very deep neural networks and further improved performance. The success of these CNN-based vision models prove the effectiveness of locality bias. Besides, these methods are all based on pyramid structure of multi-stage processing which have been widely acknowledged to improve performance. 
Later, non-local operations \cite{wang2018non} is proposed to explore global context to augment convolutional networks. It shows that researchers from this field have also realized the importance of global dependency modeling.

We believe that the experiences obtained in the CNN-based vision models, including taking advantage of locality bias through small-kernel convolutions, the pyramid structure, and exploiting global dependencies, are still valid nowadays. 
In our architecture, skip connections \cite{he2016deep} and normalization layers \cite{ioffe2015batch,ba2016layer} are also applied.  

\begin{figure*}[t]
\centering
\includegraphics[width=\textwidth]{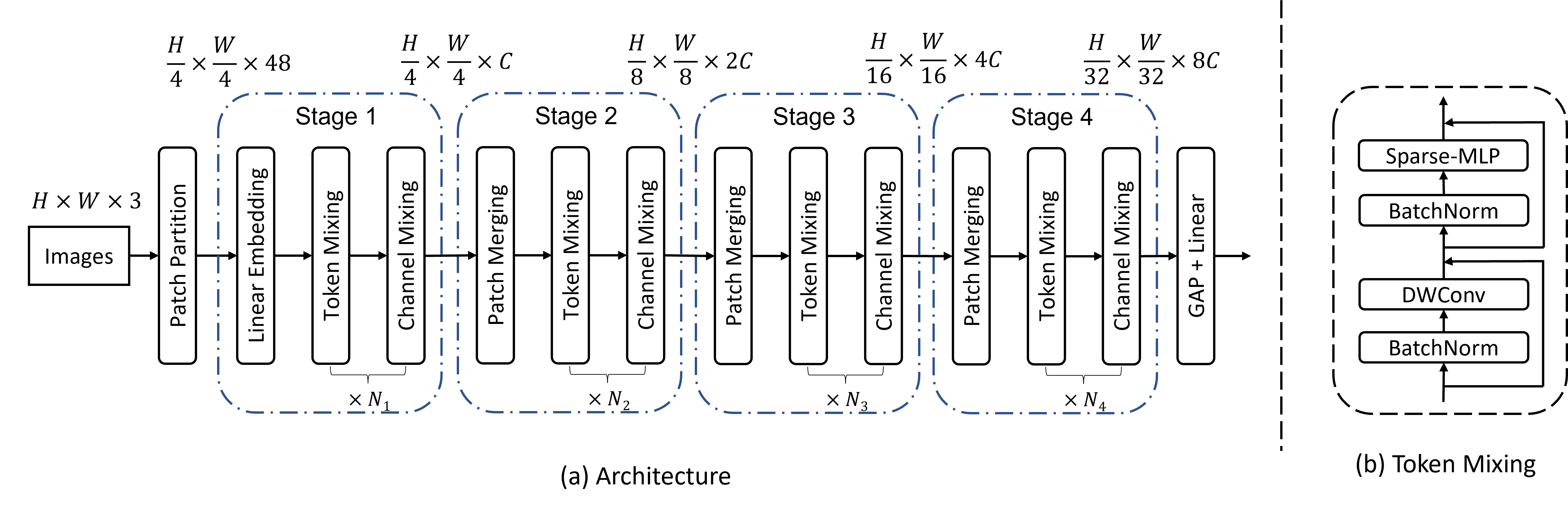} 
\caption{(a) The overall multi-stage architecture of the sMLPNet; (b) The token mixing module.}
\label{Architecture}
\end{figure*}

\subsection{Transformer-Based Vision Models}
ViT \cite{dosovitskiy2020image} is the first work to build a purely Transformer-based vision backbone. It shows that, with sufficient training data, a transformer provides better performance than a traditional CNN in vision tasks. 
DeiT \cite{touvron2021training} introduces several training strategies that allow ViT to be similarly effective using the much smaller ImageNet-1K as the training dataset. These methods demonstrate the power of global dependency modeling by self-attention. Later, some following works introduce the pyramid structure into Transformer, such as PVT \cite{wang2021pyramid}. Just as in CNN-based models, pyramid structure also improves the performance of Transformer-based models for various vision tasks. Based on the pyramid structure, Swin Transformer \cite{liu2021swin} proposes to use self-attention within each local window. This can be regarded as a use of locality bias. Following these local attention mechanism, there are also some variants such as Twins \cite{chu2021twins}, MSG-transformer \cite{fang2021msg}, GG-transformer \cite{yu2021glance} and Shuffle Transformer \cite{huang2021shuffle}. All these methods justify the importance of locality bias.

In contrast to the methods mentioned above, our method tries to model global dependencies using an attention-free mechanism. In particular, the mapping weights in self-attention modules are dynamic or data-dependent, but those in our sMLP modules are static. 

\subsection{MLP-Based Vision Models}
Our work is mostly related to MLP-based methods. MLP-Mixer \cite{tolstikhin2021mlp}, a pure MLP-like model for ImageNet classification has been proposed very recently. It uses standard dense matrix multiplications (channel-mixing feed forward networks) to aggregate information across channels. For spatial information, it flattens the 2D spatial dimensions to 1D sequence and apply another matrix multiplications (token-mixing MLPs) to aggregate information across token sequence. Despite that MLP-Mixer has achieved promising results when trained on a huge-scale dataset JFT-300M, it is not as good as its visual Transformer counterparts when trained on a medium-scale dataset such as ImageNet-1K. gMLP \cite{liu2021pay} designs a gating operation to enhance the communications between spatial locations. ResMLP \cite{touvron2021resmlp} proposes an affine transform layer which facilities stacking a huge number of MLP blocks. EA \cite{guo2021beyond} replaces the self-attention module with an external attention which is implemented by a cascade of two linear layers. The complexity in parameter and time of all these methods are quadratic with respect to the input image size when they aggregating the spatial information. Besides, there is still an accuracy gap between MLP-based methods and SOTA. We believe the main reasons are the overlook of locality bias, the absence of pyramid structure, and the over-fitting phenomenon caused by excessive number of parameters. 

In contrast to the existing MLP-based method, our method try to exploit locality bias and global dependency based on pyramid structure. Since the quadratic computational complexity of global dependency modeling by MLP is not friendly to pyramid structure, we use the proposed sparse MLP module to support the global dependency modeling based on pyramid structure.

\section{Method}
\subsection{Design Guidelines}
In this work, we intend to answer the question whether it is possible to design a high-performance network for image recognition without using self-attention modules. We want to retain some of the important design ideas used in the CNN era, and add new components inspired by Transformers. The followings are the design guidelines we try to follow:
\begin{enumerate}
    \item Adopt a similar architecture as ViT, MLP-Mixer, and Swin Transformer to ensure a fair comparison.
    \item Explicitly inject the locality bias into the network.
    \item Explore the global dependencies without the use of self-attention module.
    \item Perform multi-stage processing in a pyramid structure.
\end{enumerate}
In the next subsections, we will present the overall architecture of the proposed sMLPNet and detail the key component sMLP module. 

\subsection{Overall Architecture}

Fig.\ref{Architecture}(a) illustrates the overall architecture of our designed network. Similar to ViT, MLP-Mixer, and recent Swin transformer, an input RGB image with spatial resolution $H \times W$ is divided into non-overlapping patches by a patch partition module. We adopt a small patch size of $4 \times 4$ at the first stage of the network. Each patch is first reshaped into a 48-dimensional vector, and then mapped by a linear layer to a C-dimensional embedding. As such, the entire image is expressed as a $\frac{H}{4} \times \frac{W}{4} \times C$ tensor. 
Note that the patch size in MLP-Mixer is $16 \times 16$. With the same input image resolution, the number of tokens in our network is 16 times that in MLP-Mixer. As we know, the computational complexity of MLP grows in a quadratic way with the number of input tokens. It would be impossible for MLP-Mixer to handle such a large number of tokens if no optimization is applied. 

The entire network is comprised of four stages. Except for the first stage, which starts with a linear embedding layer, other stages start with a patch merging layer which reduces the spatial dimension by $2 \times 2$ and increases the channel dimension by $2$ times. The patch merging layer is simply implemented by a linear layer which takes the concatenated features of each $2 \times 2$ neighboring patches as input and outputs the features of the merged patch. Then, the new image tokens are passed through a token-mixing module and a channel-mixing module. These two modules do not change the data dimensions.


The token-mixing module is illustrated in Fig.\ref{Architecture}(b). In this module, we take advantage of locality bias by using a depth-wise convolution (DWConv) with kernel size 3x3. In fact, after channel processing is decomposed from spatial processing, DWConv becomes a very natural choice to explore locality. This operation is highly efficient in the sense that it contains few parameters and takes few FLOPs during inference. We also try to model global dependencies with the proposed sMLP module. The sparsity and weight-sharing nature makes sMLP less prone to over-fitting than the original MLP module. The greatly reduced computational complexity of sMLP allows us to operate at a spatial resolution of $H/4 \times W/4$ in the first stage. The details of sMLP will be presented in the next subsection. Batch normalization (BN) and skip connections are applied in a standard way in the token-mixing module. 

The channel-mixing module is implemented by an MLP or so called feed-forward network (FFN), in exactly the same way as that in MLP-Mixer. The FFN is composed of two linear layers separated by a GeLu activation. The first linear layer expands the dimension from $D$ to $\alpha D$, and the second layer reduces
the dimension from $\alpha D$ back to $D$. Here $\alpha$ is a tunable hyper-parameter and we find that $\alpha=2$ or $\alpha=3$ works the best in our network. 

\begin{figure}[t]
\centering
\includegraphics[width=0.4\textwidth]{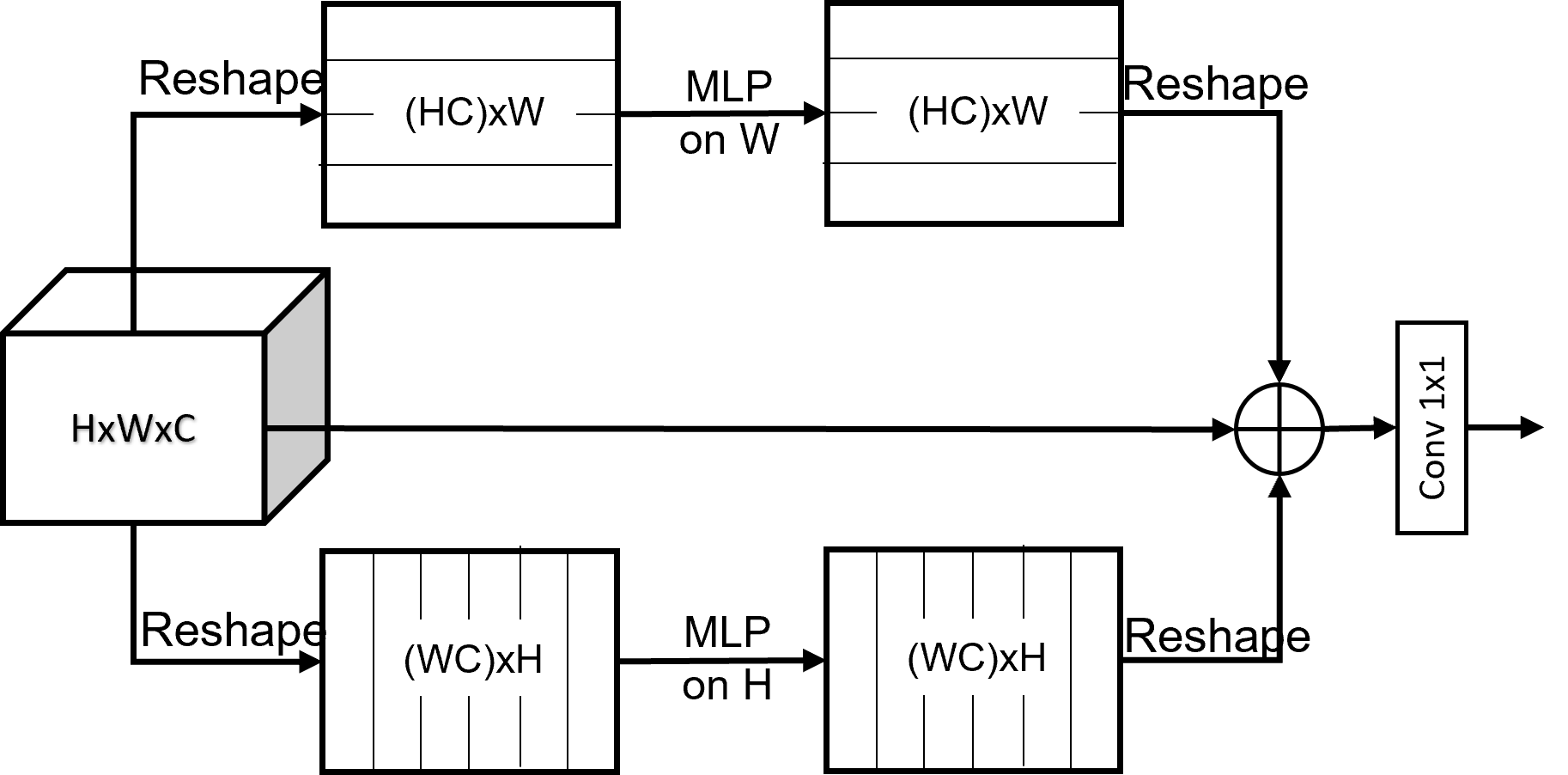} 
\caption{Structure of the proposed sMLP block. It consists of three branches: two of them are responsible for mixing information along horizontal and vertical directions respectively and the other path is the identity mapping. The output of the three branches are concat and processed by a point-wise conv to obtain the final output.}
\label{fig:sMLP}
\end{figure}

\subsection{Sparse MLP (sMLP)}
We design a sparse MLP to fix two major drawbacks of the original MLP. First, we want to reduce the number of parameters to avoid over-fitting, especially when the network is trained on moderate-sized dataset as ImageNet-1K. Second, we want to reduce the computational complexity, especially when the number of tokens is large, to enable multi-stage processing in a pyramid structure. 

In sparse MLP, we use sparse connection and weight sharing to achieve our design goal. As illustrated in Fig.\ref{fig:sMLPinteraction}, instead of interacting with all the other tokens, a token in sMLP only directly interacts with tokens on the same row or the same column. In addition, all rows and all columns can respectively share the same projection weights. Fig.\ref{fig:sMLP} shows the implementation diagram of our designed sMLP block. It consists of three paths. Besides the identity mapping shown in the middle, two other paths are responsible for mixing tokens along horizontal and vertical directions, respectively. 

Let $X^{in} \in {R}^{H \times W \times C}$ denote the collection of input tokens. In the horizontal mixing path, the data tensor is reshaped into $HC \times W$, and a linear layer with weights $W_{W} \in {R}^{W \times W}$ is applied to each of the $HC$ rows to mix information. Similar operation is applied in the vertical mixing path and the linear layer is characterized by weights $W_{H} \in {R}^{H \times H}$. Finally, the output from the three paths are fused together to produce an output data tensor which has the same dimension as the input tensor. We implement this fusion module with concatenation and a FC layer:
\begin{equation}
    X^{out} = FC(concat(X_{H},X_{W},X))
\end{equation}
The PyTorch-like pseudo code for the implementation of sMLP module can be found in Alg.~\ref{alg:algorithm}. 


This design allows each token to aggregate information across the row and the column it locates. If this module is passed twice, each token can aggregate information across the entire 2D space. In other words, sMLP efficiently obtains global receptive field although the direct connections are sparse. 

The number of parameters in one sMLP module can be computed as $H^{2} + W ^{2} +3C^{2}$, where $3C^2$ parameters are used in the fusion step. In comparison, the number of parameters in the original MLP module is $2\alpha{(HW)}^{2}$, where $\alpha$ is the expansion ratio of MLP layers which often takes value of 4. When the input image size is $224 \times 224$ and the initial patch is as small as $4 \times 4$, our sMLP module achieves about 3,000x parameter reduction when $C=80$ , which will make a huge difference in preventing the over-fitting phenomenon on moderate-sized dataset. 

The reduction of computational complexity is also prominent. Specifically, the complexity of one sMLP module is:
\begin{equation}
    \Omega(sMLP) = HWC(H+W)+3HWC^{2},
\end{equation}
and that of the token mixing part of MLP-Mixer is:
\begin{equation}
    \Omega(MLP) = 2\alpha(HW)^{2}C
\end{equation}
The product of $H$ and $W$ is the number of input tokens, denoted by $N$. It is now clear that MLP-Mixer cannot afford a high-resolution input or the pyramid processing, as the computational complexity grows with $N^2$. In contrast, the computational complexity of the proposed sMLP grows with $N\sqrt{N}$. It allows us to process a much larger $N$ and eventually enables the multi-stage processing in a pyramid structure.

\begin{algorithm}[tb]
\caption{Pseudocode of sMLP (PyTorch-like)}
\label{alg:algorithm}
\textbf{Input}: x \# input tensor of shape (H, W, C)\\
\textbf{Output}: x \# output tensor of shape (H, W, C)\\

proj\_h = nn.Linear(H,H). \\
proj\_w = nn.Linear(W,W). \\
fuse = nn.Linear(3C,C). \\

def sparse\_mlp(x): \\
\begin{algorithmic}
\STATE    x\_h = self.proj\_h(x.permute(2,1,0)).permute(2,1,0) \\
    x\_w = self.proj\_w(x.permute(0,2,1)).permute(0,2,1) \\
    x = torch.concat([x\_h,x\_w,x],dim=2) \\
    x = self.fuse(x) \\
    
    return x
\end{algorithmic}
\end{algorithm}

\subsection{Model Configurations}
We build three variants of our model, called sMLPNet-T, sMLPNet-S sMLPNet-B to match with the model size of Swin-T, Swin-S, and Swin-B, respectively. The expansion parameter in the FFN for channel mixing is $\alpha = 3$ by default. The architecture hyper-parameters of these models are: \begin{itemize}
    \item sMLPNet-T: C = 80, number of layers = [2; 8; 14; 2],
    \item sMLPNet-S: C = 96, number of layers = [2; 10; 24; 2],
    \item sMLPNet-B: C = 112, number of layers = [2; 10; 24; 2],
\end{itemize} 
where C is the number of channels of the hidden layers in the first stage. The number of layers indicate the number of times the pair of token mixing and channel mixing modules are stacked in each of the four stages. The model size and theoretical computational complexity (FLOPs) of the model variants for ImageNet image classification are listed in Table~\ref{tab:main results}.

\section{Experiments}
\subsection{Experimental Setup}
We evaluate our model based on ImageNet-1K dataset \cite{krizhevsky2012imagenet} which contains 1.2
million training images from one thousand categories and
50 thousand validation images with 50 images in each category. 

We train our model using AdamW \cite{loshchilov2018decoupled} with weight decay 0.05 and a batch size of 1024. We use a linear warm up and cosine decay. The initial learning rate is 1e-3 and gradually drops to 1e-5 in 300 epochs. We also use label smoothing \cite{szegedy2016rethinking} and DropPath \cite{larsson2016fractalnet}. DropPath rates for our tiny, small, and base models are 0, 0.2, and 0.3, respectively. For data augmentation methods, we use RandAug \cite{cubuk2020randaugment}, repeated augmentation \cite{hoffer2020augment}, MixUp \cite{zhang2018mixup}, and CutMix \cite{zhong2020random}. All training is conducted with 8 NVIDIA Tesla V100 GPU cards.

\subsection{Ablation Study}
We carry out ablation studies on all three variants of the sMLPNet. Due to space limit, we will only present the numerical results on one variant in each ablation experiment. The trend on other variants is the same if not otherwise stated. Similarly, we also conducted experiments on both choices of the expansion parameter, $\alpha=2$ and $\alpha=3$, in the FFN. We randomly choose one setting to report the numerical results. In order to differentiate, we append a * to the model name when $\alpha=2$.

\subsubsection{Local and Global Modeling.}
\begin{table}[t]
\centering
\begin{tabular}{c|c c c c}
    \hline
    sMLPNet-T* & Param(M) & FLOPs(B) & Top-1(\%) \\
    \hline
    Local+Global  &19.2 &4.0 &81.3  \\
    Global only   &19.1 &3.9 &80.6  \\
    Local only    &22.5 &4.4 &80.7  \\
    \hline
\end{tabular}
\caption{Ablation study on the effects of local and global modeling using the tiny model ($\alpha=2$).}
\label{tab:local and global}
\end{table}
In sMLPNet, we use depth-wise convolution (DWConv) to model locality and sMLP to model global dependencies. In order to verify the need to model both types of dependencies, we remove either DWConv or sMLP to check how the top-1 accuracy changes. The base model is the tiny version sMLPNet-T with FFN expansion parameter $\alpha=2$. The top-1 accuracy achieved by this base model is 81.3\%. 

As we have mentioned, the DWConv operation is extremely lightweight. When we remove it from sMLPNet, the model size only changes from 19.2M to 19.1M and the FLOPs only decrease by 0.1B, as shown in Table \ref{tab:local and global}. However, the image recognition accuracy significantly drops to 80.6\%. This clearly shows that DWConv is a very efficient way to model local dependencies and that a vision model should take advantage of the inductive bias on locality. 

We then remove the sMLP module to evaluate the performance of a network with local modeling only. Since sMLP is quite heavy, in order to ensure a fair comparison, we increase the number of channels from 80 to 112 to make the model size and FLOPs roughly comparable to the base model. From Table \ref{tab:local and global}, we can see that the local only version only achieves an accuracy of 80.7, although the model being tested is slightly larger than the base model. This experiment confirms that both local and global modeling are important in sMLPNet. 

\begin{table}[t]
\centering
\begin{tabular}{c c c c|c c c}
    \hline
    S1 & S2 & S3 & S4 & Param(M) & FLOPs(B) & Acc.(\%) \\
    \hline
    \checkmark & \checkmark & \checkmark & \checkmark &65.9  &14.0 &83.4  \\
    & \checkmark & \checkmark & \checkmark &65.8  &13.7  &83.2  \\
    & & \checkmark & \checkmark &64.3  &12.4 &83.0  \\
    & & & \checkmark &49.9  &9.5  &82.2  \\
    & & & \textcolor{white}{\checkmark} &45.1  &9.3  &82.0  \\
    \hline 
\end{tabular}
\caption{Abation study on the effects of sMLP using sMLPNet-B ($\alpha=3$) as the base model. We remove the sMLP block from the beginning of the network and evaluate the top-1 accuracy. A check mark in the corresponding stage (S1, S2, S3, and S4) means the use of sMLP module.}
\label{tab:Local & global}
\end{table}

After verifying the overall validity of the sMLP module, we further attempt to explore its role in different stages of the network. For this experiment, we use sMLPNet-B as the base model. It has 65.9M parameters and 14.0B FLOPs. The top-1 accuracy is 83.4\%. Then, we start to remove sMLP from stage 1 until stage 4. After removing sMLP from an entire stage, we train the model and report the image recognition results, which are listed in Table.\ref{tab:Local & global}. Note that the first row in the table corresponds to the full base model and the last row is the local only version of sMLPNet-B.

It is not quite surprising to see that the top-1 accuracy decreases as we remove sMLP module from more stages. The decrease in accuracy is roughly proportional to the reduction of model size and FLOPs. The largest accuracy drop happens when we remove sMLP from stage 3. The accuracy decreases from 83.0\% to 82.2\% while the model size is reduced from 64.3M to 49.9M and the FLOPs is reduced from 12.4B to 9.5B. Note that, after removing sMLP from the first three stages, the resulting model is obviously inferior to the sMLPNet-S based model which obtains 83.1\% accuracy at a model size of 48.8M. This result emphasizes the necessity of global dependency modeling in early stages of the network. 

One may wonder why sMLP module brings the most number of parameters and computational cost in stage 3. Besides the fact that stage 3 has the most number of layers, it is mainly caused by the expansion of the fusion module, whose parameter size and computational complexity both grow with the square of the number of channels. It is natural to ask whether there exists a more efficient fusion method than an FC layer. 


\subsubsection{Fusion in sMLP.}
\begin{table}[t]
\centering
\begin{tabular}{c|c c c c}
    \hline
    & Param(M) & FLOPs(B) & Top-1(\%) \\
    \hline
    sMLPNet-S     &48.6 &10.3 &83.1  \\
    \hline
    Sum    &33.2 &7.0 & 81.5  \\
    Weighted sum    &33.3 &7.0 &81.8  \\
    \hline
    sMLPNet-T & 24.1 & 5.0 & 81.9 \\
    \hline
\end{tabular}
\caption{Comparison of different fusion methods. Base model is sMLPNet-S ($\alpha=3$) which uses an FC layer for data fusion. Sum and weighted sum are two alternative fusion methods.}
\label{tab:Fusion method}
\end{table}
We try two light-weight operations as the alternatives to the fusion method in sMLPNet. One is element-wise addition, which is parameter-free and costs few FLOPs in run time. The other is weighted sum. Specifically, each channel is multiplied by a learnable weight before addition. 
The number of parameters in weighted sum operation is also very small and the computational cost almost negligible. 

We use sMLPNet-S as the base model and the experimental results are shown in Table \ref{tab:Fusion method}. Besides the fusion method, we do not change any other modules in the network, including the number of channels and the number of layers in each stage. Compared to baseline, which has 48.6M parameters and 10.3B FLOPs, the two alternative fusion methods bring much fewer parameters and FLOPs. But the image recognition accuracy also drops from 83.1\% to 81.5\% and 81.8\%. While it is understandable that the accuracy decreases with the decreased size of model, we list the performance of sMLPNet-T as a reference. sMLPNet-T uses the default fusion method and it achieves 81.9\% accuracy at a model size of 24.1M. This confirms that using concatenation and an FC layer as the fusion module achieves the best model size and accuracy trade-off among the different fusion methods we evaluated. 

\subsubsection{Branches in sMLP.}
\begin{table}[t]
\centering
\begin{tabular}{c|c c}
    \hline
    & Parallel & Sequential \\
    \hline
    w Identity   &81.3 &81.1 \\
    w/o Identity &80.9 &80.6  \\
    \hline
\end{tabular}
\caption{Ablation study on the design of the branches in the sMLP module. We use sMLPNet-T* ($\alpha=2$) as the base model (parallel connection with the identity mapping).}
\label{tab:Parallel versus Sequential}
\end{table}

In the default setting of sMLP, we use three parallel branches for horizontal processing, vertical processing, and identity mapping, respectively. In this experiment, we try a different fashion to connect the horizontal processing and the vertical processing. We also verify the effectiveness of the identity mapping. 

The base model for this set of experiments is sMLPNet-T with expansion parameter $\alpha=2$. The two choices to connect horizontal and vertical processing are parallel, as in the default setting, and sequential. We combine these two choices with the choice to use or not use the identity mapping to create four settings. The performance of these four settings are shown in Table \ref{tab:Parallel versus Sequential}. 

If we compare results within each column, we can tell that using identity mapping always brings better results. The improvement is in the range of 0.4\% and 0.5\%, which is considered significant. If we compare results within each row, we can tell that parallel processing is always better than sequential processing. Note that sequential mode can be implemented by horizontal first or vertical first. We tried both settings and got exactly the same results. As such, we validate our selection of parallel mode with identity mapping in the final design of sMLP block.

\subsubsection{Multi-Stage Processing in Pyramid Structure.}
\begin{table}[t]
\centering
\begin{tabular}{c|c c c c}
    \hline
    & Param(M) & FLOPs(B) & Top-1(\%) \\
    \hline
    sMLPNet-T*    &19.2 &4.0 &81.3  \\
    Multi-stage MLP     &22.7 &4.2 &77.8  \\
    Single-stage MLP    &30.4 &6.5 &76.8  \\
    \hline
\end{tabular}
\caption{Ablation study on the effects of multi-stage architecture. Multi-stage MLP: sMLP at stage 1 is replaced by depth-wise conv and the sMLP at stage 2,3,4 is replaced by MLP. Single-stage MLP: all of the sMLP is replaced to MLP and multi-stage is replaced to singe-stage with patch size equal to $16\times16$.}
\label{tab:multi-stage}
\end{table}

Enabling multi-stage processing in a pyramid structure is one of the major objectives of our work. Such a design choice has been validated in CNNs for numerous network backbones and downstream tasks. While it is intuitive to stick to this choice when global dependency modeling is added to the network, we still need to verify it through numerical evidence. 

In this experiment, we are actually comparing the single-stage version and the multi-stage versions of MLP networks. But we built these two models from a base sMLPNet model. Specifically, we take a tiny sMLPNet model ($\alpha=2$) and replace all the sMLP blocks in stage 2, 3, and 4 with the normal MLP blocks. The sMLP blocks in stage 1 is replaced by DWConv, as MLP blocks are too heavy to be used in stage 1. This is referred to as the multi-stage MLP model in Table \ref{tab:multi-stage}. We can see that the top-1 accuracy of this model is only 77.8\%, with 3.5\% performance loss with respect to the base sMLPNet-T model. 

Then, we flatten the spatial resolutions of all stages and use an initial patch size of $16 \times 16$ to build the single-stage MLP model. Table \ref{tab:multi-stage} shows that it achieves an even lower top-1 accuracy of 76.8\% on ImageNet-1K dataset. This is consistent with the performance reported by MLP-Mixer when it is trained on ImageNet-1K. This experiment clearly shows the advantage of multi-stage processing. In addition, it demonstrates the huge gain we have achieved over the straightforward design of an MLP network. 

\subsection{Comparison with State-of-the-Art}

\begin{table}[t]
\centering
\begin{tabular}{c|c c c}
    \hline
    Model& Params(M)& FLOPs(B) & Top-1(\%) \\
    \cline{2-4}
    & \multicolumn{3}{|c} {ConvNets} \\
    \hline
    ResNet-50 & 25.6 & 4.1 & 76.2 \\
    ResNet-152 & 60.2 & 11.5 & 78.3 \\
    RegNetY-4GF & 21 & 4.0 & 80.0 \\
    RegNetY-8GF & 39.2 & 8.0 & 81.7 \\
    RegNetY-16GF & 83.6 & 15.9 & 82.9 \\
    \hline
    & \multicolumn{3}{|c} {Transformers} \\
    \hline
    ViT-B/16 & 86.4 & 17.6 & 79.7 \\
    DeiT-S/16 & 22 & 4.6 & 79.8 \\
    DeiT-B/16 & 86.4 & 17.6 & 81.8 \\
    DeepVit-S & 27 & 6.2 & 81.4 \\
    DeepVit-L & 55 & 12.5 & 82.2 \\
    Swin-T & 29 & 4.5 & 81.3 \\
    Swin-S & 50 & 8.7 & 83.0 \\
    Swin-B & 88 & 15.4 & 83.5 \\
    \hline
    & \multicolumn{3}{|c} {MLP-like} \\
    \hline
    Mixer-B/16 & 59 & 12.7 & 76.4 \\
    Mixer-L/16 & 207 & 44.8 & 71.8 \\
    ResMLP-12& 15 & 3.0 & 76.6 \\
    ResMLP-24& 30 & 6.0 & 79.4 \\
    ResMLP-36 & 45 & 8.9 & 79.7 \\
    gMLP-S& 20 & 4.5 & 79.4 \\
    gMLP-B& 73 & 15.8 & 81.6 \\
    \hline
    sMLPNet-T* (ours) &19.2 &4.0 &81.3 \\
    sMLPNet-T (ours) &24.1 &5.0 &81.9 \\
    sMLPNet-S (ours) &48.6 &10.3 &83.1  \\
    sMLPNet-B (ours) &65.9 &14.0 &83.4  \\
    \hline
\end{tabular}
\caption{Comparing the proposed sMLPNet with state-of-the-art vision models. The default expansion parameter in the FFN of sMLPNet is $\alpha=3$. sMLPNet-T* uses $\alpha=2$. All models are trained on ImageNet-1K benchmark without extra data. The resolution of the input image is $224 \times 224$ for all the models.}
\label{tab:main results}
\end{table}

We have validated the various design choices in the proposed sMLPNet. Next, we show how this attention-free network compares with state-of-the-art vision models on the image recognition task. All the models we refer to are trained on ImageNet-1K benchmark only. For fairness, all of them use an input image size of $224 \times 224$. The results are summarized in Table~\ref{tab:main results}. 

We group the existing methods into three categories, namely CNN-based, Transformer-based, and MLP-like models. Some of the designs have three different model sizes while some others have two. The model size is indicated by the suffix of T (tiny), S (small), B (base), and L (large). We also list the theoretical model size and FLOPs in the table for the comparison across different designs. 

In the CNN-based vision models, the RegNetY \cite{radosavovic2020designing} series have a large advantage over the classic ResNet series. Transformer-based models \cite{touvron2021training, zhou2021deepvit, liu2021swin}, except the very first ViT model \cite{dosovitskiy2020image}, perform on par with RegNetY series. But most MLP-like models \cite{touvron2021resmlp, liu2021pay} only achieve similar performance as ResNet series. Besides, serious over-fitting phenomenon is observed in Mixer-L model. Its accuracy dramatically drops from 76.4\% of the Mixer-B model to 71.8\%. It explains from the opposite side why sMLPNet achieves good performance by the reduction of parameters.

Among these existing models, Swin Transformer performs the best. Our designed model, despite the fact that it belongs to the MLP-like category, performs on par with or even better than Swin Transformer. In particular, sMLPNet-T achieves 81.9\% top-1 accuracy, which is the highest among the existing models with FLOPs fewer than 5B. 

The performance of sMLPNet-B is also very impressive. It achieves the same top-1 accuracy as Swin-B, but the model size is 25\% smaller (65.9M vs. 88M) and the FLOPs are nearly 10\% fewer (14.0B vs. 15.4B). Remarkably, there is no sign of over-fitting, which is the main problem that plagues the MLP-like methods, when the model size grows to nearly 66M. This shows that an attention-free model could attain SOTA performance, and the attention mechanism might not be the secret weapon in the top-performing Transformer-based models. 





\section{Conclusion and Discussion}
In this paper, we have built an MLP-like architecture for visual recognition based on the novel sMLP block. The sMLP block we have proposed in this work is featured by sparse connection and weight sharing. By separately aggregating information along axial directions, sMLP avoids the quadratic model size and quadratic computational complexity of conventional MLP. Experimental results have shown that this has greatly pushed up the performance boundary of MLP-like vision models.

We notice that some concurrent Transformer-based models, such as CSwin \cite{dong2021cswin}, have obtained an even higher accuracy than sMLPNet. For example, 84.2\% top-1 accuracy is achieved with a model of 78M parameters. But this does not affect the value of our work, which challenges the necessity of the self-attention mechanism by demonstrating how well an attention-free network could perform. 

That being said, we have to admit that MLP-like architecture has its inherent limitation. Due to the fixed nature of FC layers, MLP-like models cannot be easily adapted to process input images with arbitrary resolutions. This makes MLP-like models hard to be applied in some important down-stream tasks such as object detection and semantic segmentation. Again, it is an inherent drawback of MLP-like architecture and is beyond the scope of our work. In the future, we plan to investigate the possibility to build attention-free versatile networks.


\appendix
\label{sec:reference_examples}

\nobibliography*
\bibliography{aaai22}

\end{document}